\documentclass[ 5p]{elsarticle}

\usepackage{lineno,hyperref,tikz,bm,bbold,verbatim,verbatimbox,multicol,multirow}
\modulolinenumbers[200]
\usetikzlibrary{mindmap,backgrounds, snakes, shapes, positioning}
\journal{Computer Methods and Programs in Biomedicine}

\newcommand\ind{I}
\newcommand\code[1]{{\normalfont\fontfamily{cmvtt}\selectfont #1}}
\newcommand\hist[1]{\overline{#1}}
\newcommand\SL{\%SuperLearner}

\begin{document}

\begin{frontmatter}

\title{Super learning in the SAS system}

\author{Alexander P Keil\fnref{myfootnote}, 
Daniel Westreich\fnref{myfootnote}, 
Jessie K Edwards\fnref{myfootnote}, 
Stephen R Cole\fnref{myfootnote}, 
}
\address{2102-A McGavran Greenberg CB 7435, Chapel Hill, NC}
\fntext[myfootnote]{Department of Epidemiology, Gillings School of Global Public Health, University of North Carolina at Chapel Hill}

\begin{abstract}

\textbf{Background and objective}
Stacking is an ensemble machine learning method that averages predictions from multiple other algorithms, such as generalized linear models and regression trees.
An implementation of stacking, called super learning, has been developed as a general approach to supervised learning and has seen frequent usage, in part due to the availability of an R package.
We develop super learning in the SAS software system using a new macro, and demonstrate its performance relative to the R package.

\textbf{Methods}
Following previous work using the R SuperLearner package we assess the performance of super learning in a number of domains.
We compare the R package with the new SAS macro in a small set of simulations assessing curve fitting in a predictive model as well in a set of 14 publicly available datasets to assess cross-validated accuracy.

\textbf{Results}
Across the simulated data and the publicly available data, the SAS macro performed similarly to the R package, despite a different set of potential algorithms available natively in R and SAS.

\textbf{Conclusions}
Our super learner macro performs as well as the R package at a number of tasks.
Further, by extending the macro to include the use of R packages, the macro can leverage both the robust, enterprise oriented procedures in SAS and the nimble, cutting edge packages in R.
In the spirit of ensemble learning, this macro extends the potential library of algorithms beyond a single software system and provides a simple avenue into machine learning in SAS.

\end{abstract}

\end{frontmatter}



\section{Introduction}
Supervised machine learning is a generic term for algorithms that utilize measured values of some outcome ("supervisor") to build models for predicting future values of the outcome, given new inputs. 
Supervised machine learning is emerging as an essential tool for prediction and causal inference in biomedicine.
Ensemble machine learning algorithms combine multiple algorithms into a single learner that can improve prediction characteristics such as classification accuracy or prediction error.
One ensemble machine learning method, referred to as stacking, is an approach to combining an arbitrary set of learning algorithms, including other ensemble methods\cite{RN6165,RN6121}.
A recent approach to stacking, referred to as super learning, has demonstrated theoretical and practical properties that make it a sound default framework for prediction \cite{RN567,RN566}.
One of the practical properties is the relative ease of implementation that has led to the development of several software packages for super learning.
In turn, the availability of software has made the approach relatively simple to use in research \cite{RN6166}.

Existing implementations of Super Learner include an R package \cite{slpackage}, maintained by the developers of the super learning algorithm, and unofficial, open source releases in Python \cite{Lendle:2014aa,Keil:2017aa}, and a small open source version in SAS \cite{Brooks:2016aa}.
The existing SAS implementation is not under active development and has a limited library of algorithms.

We demonstrate usage of super learning in the SAS system by introducing a new SAS macro, \SL{} (\url{https://cirl-unc.github.io/SuperLearnerMacro}).
This macro improves on existing software by providing a general approach to super learning with an extensive existing library that is easily extensible by the user to incorporate new learners, including algorithms from the R programming language.
We demonstrate the use of this macro using one simulated example and one example using multiple real-world datasets, and we compare performance with existing implementations.
These examples closely follow the analyses of Polley and van der Laan, which demonstrated the R SuperLearner package \cite{RN566,slpackage}.

\section{Methods: supervised, super learning \label{sec:learning}}
We first provide a brief review of supervised machine learning in the context of epidemiologic data, and then describe the super learning algorithm.
\subsection{Supervised learning}
Suppose one is interested in learning about how lung cancer mortality rates vary according to age, smoking, and radon exposures in a population of uranium miners from Colorado in the 1950s.
One can frame such learning in terms of a causal inference problem (e.g.
what would be the change in the lung cancer mortality rates if one could eliminate smoking among the miners?) or in terms of a prediction problem (e.g.
what is the expected lung cancer mortality rate among a non-smoking 70 year old former miner who was exposed at the Mining Safety and Health Administration occupational exposure limit from ages 20 to 65?).
Supervised machine learning is one way to use the inputs $\bm{X}$ (smoking, age, radon exposure) and outputs $Y$ (lung cancer mortality) as a way to describe patterns in how $\bm{X}$ relates to $Y$. Like any purely statistical approach, machine learning is agnostic to whether these relations are causal or associational/predictive.

We can describe the relation between these variables through a function $f(\bm{x};\bm{\beta},S)$ that yields the average lung cancer mortality rate $\hat{Y}$ for a given pattern of smoking and radon exposure at a given age within the context of our study sample $S$.
$f(\bm{x};\bm{\beta},S)$ is used to estimate $E(Y|\bm X)$, the conditional expectation of $Y$ given $\bm X$.
The parameters $\bm{\beta}$ determine the shape of the function that relates inputs to outputs.
For example $\bm{\beta}$ could represent log-rate ratios if our function is a Poisson regression model or it could represent the assigned values of $\hat{Y}$ within nodes of classification and regression trees.

Using the study data and a some function (e.g. logistic model, regression tree), one ``trains'' the parameters $\bm{\beta}$ of that function by finding the unique values of $\bm{\beta}=\bm{\hat\beta}$ that minimize some estimated expected loss function, given by $\hat{E}[\hat{L}(\hat{Y},Y)]$.
Most readers will be familiar with some common loss functions such as the negative log-likelihood (as in maximum likelihood estimation) or squared-error loss $\hat{E}[\hat{L}(\hat{Y},Y)] = N^{-1}\sum_i(\hat{\epsilon}_i)^2$ where $\hat{\epsilon}_i =Y_i-f(\bm{x}_i;\bm{\hat\beta},S)=Y_i-\hat Y_i$ (as in ordinary least-squares).
In this context, supervised machine learning is the act of training the parameters  $\bm{\beta}$  such that the function $f(\bm{x};\bm{\beta},S)$ estimates $E(Y|X)$ and thus carries information about how $\bm{X}$ relates to the expected $Y$ in the study sample.  
We refer to such an estimate as a "prediction."
Parameters of a model are ``learned'' from the data by minimizing some distance measure between observed values of $Y$ and model predictions of $Y$ (given $\bm{X}$), which is just to say that the goal of supervised learning is to make accurate predictions, given values of $\bm{X}$. We note here for generality that we might also chose $f(\bm{x};\bm{\beta},S)$ to estimate other conditional quantities such as the conditional median of $Y$, given $\bm X$, but for clarity we focus on $E(Y|X)$.

Throughout, we assume these data are independently and identically distributed (i.i.d.), implying that the function $f(\bm{x};\bm{\beta},S)$ can take as input some individual's covariates $\bm{x}_i$ and output an individual level prediction $\hat{y}_i$ without considering covariate values of other individuals and allowing that the same function applies to each unit in population $S$.

\subsection{Super learning}
For a given predictive or inferential problem, we often have many available choices of functions, or learners, in order to make predictions, and there is often little \emph{a priori} information to select one particular learner. For example, commonly epidemiologists might choose between a logistic regression model and a log-binomial model for data with a dichotomous outcome.
However, results may crucially depend on this choice.
Super learning is a way of combining multiple learners using cross-validation as a way to reduce dependency of the results on the choice of learner (for example, generating predictions that strike a compromise between logistic and log-binomial models).
Precise, theoretic descriptions of super learning are given in \cite{RN567} and \cite{RN566}, but we review basic principles here.

Let $M$ be the number of learners in the library (set) of learners, and index each learner as $f_m(\bm{x};\bm{\beta},S)$  for $m \in 1,\ldots,M$.
We denote predictions from the $M$ learners by the vector $\hat{\bm{Y}}_{\hist{m}} \equiv (Y_1, \ldots, Y_m)$.
For example, $M$ could equal 3 and our library could contain a generalized linear model (\code{glm}), and a regression tree (\code{tree}), and $\hat{\bm{Y}}_{\hist{m}} \equiv (\hat Y_{glm}, \hat Y_{tree})$.
The super learner prediction $\hat{Y}_{sl}$ is given as a combination of the predictions from the $M$ leaners, which can be expressed as in equations \ref{eqn:level0} and \ref{eqn:sleqn}.

\begin{eqnarray}
\mbox{Level-0: } \hat{Y}_{m} = &f_m(\bm{X};\bm{\beta}_m,S) \mbox{ for }m \in 1,\ldots,M        \label{eqn:level0}\\
\mbox{Level-1: } \hat{Y}_{sl} = &f_{sl}(\hat{\bm{Y}}_{\hist{m}};\bm{\alpha},S)                   \label{eqn:sleqn}
\end{eqnarray}

We adopt Wolpert's terminology and refer to the functions $f_m$ as ``level-0'' models, which are  regression models for the observed $Y$ on covariates $\bm{X}$ indexed by parameters $\bm{\beta}_m$. We refer to the function $f_{sl}$ as a ``level-1'' model in which the observed $Y$ is regressed on the set of predictions $\hat{\bm{Y}}_{\hist{m}}$ using a model indexed by parameters $\bm{\alpha}$ \cite{RN6165}. Typically, we constrain $\bm{\alpha}$ to be non-negative and sum to 1.0, in which case the super learner prediction is a simple weighted average of predictions from each algorithm, with the weights given by $\bm{\alpha}$.


As written, the models in equations \ref{eqn:level0} and \ref{eqn:sleqn} are ``trained'' in that $\bm{\alpha}$ or $\bm{\beta}_m$ are already known.
In most problems we will not know the parameters so they must be estimated, or trained.
In combination with the level-1 model, $V$-fold cross-validation (see below) is used as a way to estimate values of the parameters that yield the best out-of-sample predictive performance, given the data and the library of learners, by minimizing the expected $V$-fold cross-validated loss.
The goal of super learner is to estimate parameters from the data that will yield a model with accurate predictions in other datasets on which the parameters have not been trained, but are drawn from the same distribution as the training data.

$V$-fold cross-validation proceeds as follows:
\begin{description}
\item[Partitioning] Begin by partitioning the data into $V\leq N$ equally sized folds ($N$ is the size of the study sample).
Typically, $V$ = 10 or 20.
We denote $v \in 1,\ldots,V$ as the $v$th fold, and  $\neg v$ denotes the remaining $V-1$ folds. 

\item[Training] For each fold $v$, we train the parameters on the remaining folds (denoted by  $\bm{\hat\beta}_{m\neg v}$), and then using those values to make predictions of $Y$, given the values of $\bm X$ in the $v$th fold of the study sample, denoted $S_v.$  This yields $V$ sets of trained parameters.


\item[Predicting] For each fold, $v$, we make predictions based on the trained parameters  $\bm{\hat\beta}_{m\neg v}$. This yields ``cross-validated predictions'' for each fold, which are denoted as $\hat{Y}_{mv} = f_m(\bm{x_v};\bm{\hat\beta}_{m\neg v},S_v)$. This notation emphasizes that, while the predictions are for study sample members in fold $v$ of the data, the parameters were trained only using data from other folds. This process results in one prediction for each member of the study sample.

\item[Combining] Cross-validated predictions are then combined across folds. The cross-validated predictions for learner $m$ are denoted as $\hat{Y}_{{m}}^{cv} \equiv (\hat{Y}_{m1}^T,\ldots,\hat{Y}_{mV}^T)^T$, where $T$ means ``transpose.'' If $Y$ is a vector of length $N$, then $\hat{Y}_m^{cv}$ will also be a vector of length $N$. This combined set of predictions is used to estimate $\bm{\alpha}$, which represent the contribution of each learner to the super learner prediction.
\end{description}
The coefficients $\bm{\alpha}$ are estimated in a model of the form $f_{sl}(\hat{\bm{Y}}^{cv}_{\hist{m}};\bm{\hat \alpha},S)$, which is essentially a fitted regression model identical to the level-1 model above, but cross-validated predictions from each of the level-0 models are used in place of the ``true'' predictions.

The super learning algorithm estimates the parameters $\bm{\hat\alpha}$ that minimize the estimated expected cross-validated loss function $\hat{E}[\hat{L}(f_{sl}(\hat{\bm{Y}}^{cv}_{\hist{m}};\bm{\hat \alpha},S),Y)]$.
The final super-learner prediction is made using predictions from the level-1 model with inputs $\hat{\bm{Y}}_{\hist{m}}$ estimated in the full study sample and trained parameters $\bm{\hat\alpha}$. Computationally, this implies that for $V$-fold cross validation, each algorithm will be trained $V+1$ times.

\paragraph{Historical note} The backbone for  super learning (or ``stacked generalization'') was laid out by Wolpert \cite{RN6165} and developed further by Breiman \cite{RN6121} to improve the finite sample performance in a simple class of linear level-1 models.
The algorithm given in equations \ref{eqn:level0} and \ref{eqn:sleqn}, and underlying theory, was generalized to arbitrary functions $f_m$ and $f_{sl}$ by van der Laan et al., who allowed that  $f_{sl}$ could be, for example, a penalized regression model or a random forest \cite{RN567}; use of the class of algorithms under this generalization was termed ``super learning.'' In practice, however, modern super learning algorithms are relatively unchanged from stacking algorithms in place by the late 1990s, which rely on parametric linear level-1 models in which the parameters $\bm{\alpha}$ form a convex combination (i.e.
$\sum_m \alpha_m = 1; \alpha_m>0 \mbox{ for }m \in 1,\ldots,M $ ); thus, super learner predictions can often be expressed as weighted combinations of a set of other machine learning algorithms.

\section{Comparing the \SL{} macro with R implementation.}

In the remainder of the manuscript, we give three example applications of super learning using the \SL{} macro, including one simulation study and two examples with real-world data.
The \SL{} macro is available from the github page of the author, and the most current version of the macro can be enabled in SAS by including the following at the top of a SAS program: 
\begin{verbatim}
FILENAME SASSprLrnr URL "https://git.io/fhFQd";
%INCLUDE SASSprLrnr;
\end{verbatim}



\subsection{Example 1: Simulation studies \label{sec:sims}}
Polley and van der Laan (2010) demonstrated the performance of super learner in a simple simulation problem.
This simulation involved learning a regression function characterizing the mean of a continuous variable $Y$ as a function of a single continuous predictor $X$, given as a uniform random variable with min=-4, max=4.
$Y$ was generated under four different scenarios, given by:
\begin{description}
  \item[Sim 1] $y_i = -2\ind(x_i<-3) + 2.55\ind(x_i>-2)-2\ind(x_i>0)+4\ind(x_i>2)-\ind(x_i>3) + \epsilon_i$
  \item[Sim 2] $y_i = 6+0.4x_i - 0.36x_i^2 + 0.005x_i^3 + \epsilon_i$
  \item[Sim 3] $y_i = 2.83\sin(\pi/2x_i) + \epsilon_i$
  \item[Sim 4] $y_i = 4\sin(3\pi x_i) \ind(x_i>0) + \epsilon_i$
\end{description}

where $\ind(condition)$ is the indicator function taking on value 1 if the logical condition is true, and 0 otherwise, and $\epsilon_i$ are standard-normally distributed error terms. This set of simulations quantifies how well a learning algorithm with parameters estimated in a training set of N=100 could predict outcomes in a test data set of size 10,000 (i.e. they assess out-of-sample predictive accuracy).
The metric used is $R^2$ which we estimate by 

$$\hat{R}^2 = 1-\frac{\sum_i\left(Y_i-\hat{Y}_i\right)^2}{\sum_i\left(Y_i-\hist{Y}\right)^2}$$

where $\hist{Y}$ is the sample mean of $Y$. The optimal value of $\hat{R}^2$ (the expected value under the true parametric model) is 0.80 for all four simulations, where $\hat{R}^2$ is estimated in the test data.
Average estimated $\hat{R}^2>0.8$ (using in-sample metrics) would imply over-fit: fitting or explaining the random error term from the data $\epsilon_i$ using $x_i$, which are \emph{a priori} independent. 
Over-fit generally results in poor predictions outside of the data in hand, which we could observe as mean $\hat{R}^2<<0.8$ in the validation data (though this could equally imply under-fit).
The simulations were each repeated 100 times and $\hat{R}^2$ was calculated for super learner and each learner in the super learner library. For each algorithm, and for super learner, we estimated the mean and interquartile range of the $\hat{R}^2$ estimates across the 100 simulations.

To compare performance of our SAS macro with an existing implementation of super learner in R, we slightly modified Polley and van der Laan's original simulation analysis to create a super learner library with algorithms that were available in both SAS and R.
Our super learner library contained the following: linear regression with only main terms (\code{glm}) or including all first order interaction terms  \code{glm + intx}), random forest (\code{rf}\cite{RN3889}), bootstrap aggregation of trees (\code{bagging}\cite{Peters:2017aa}), generalized additive models (\code{g.a.m.}\cite{Hastie:2018aa}), gradient boosting (\code{boosting}\cite{Chen:2018aa}), neural networks (\code{neural net}\cite{Venables:2002aa}), multivarate adaptive regression splines (\code{m.a.r.s.}\cite{Kooperberg:2015aa}), Bayesian additive regression trees (\code{b.a.r.t.}\cite{Chipman:2010aa}), and local polynomial regression (\code{loess}\cite{Cleveland:1991aa}).
Variations of some of these algorithms were added to the super learner library: \code{bagging} algorithms with complexity parameters set to 0.0, 0.01, and 0.1 were used, as well as one with a mean split size (ms) of 5; \code{g.a.m.} algorithms were created using splines with 2, 3, or 4 degrees of freedom; \code{neural net} algorithms were created with 2,3,4, or 5 hidden nodes; finally \code{loess} algorithms were created with smoothing parameters set to 0.75, 0.5, 0.25, or 0.1.

\begin{figure*}
  \includegraphics[width=\textwidth]{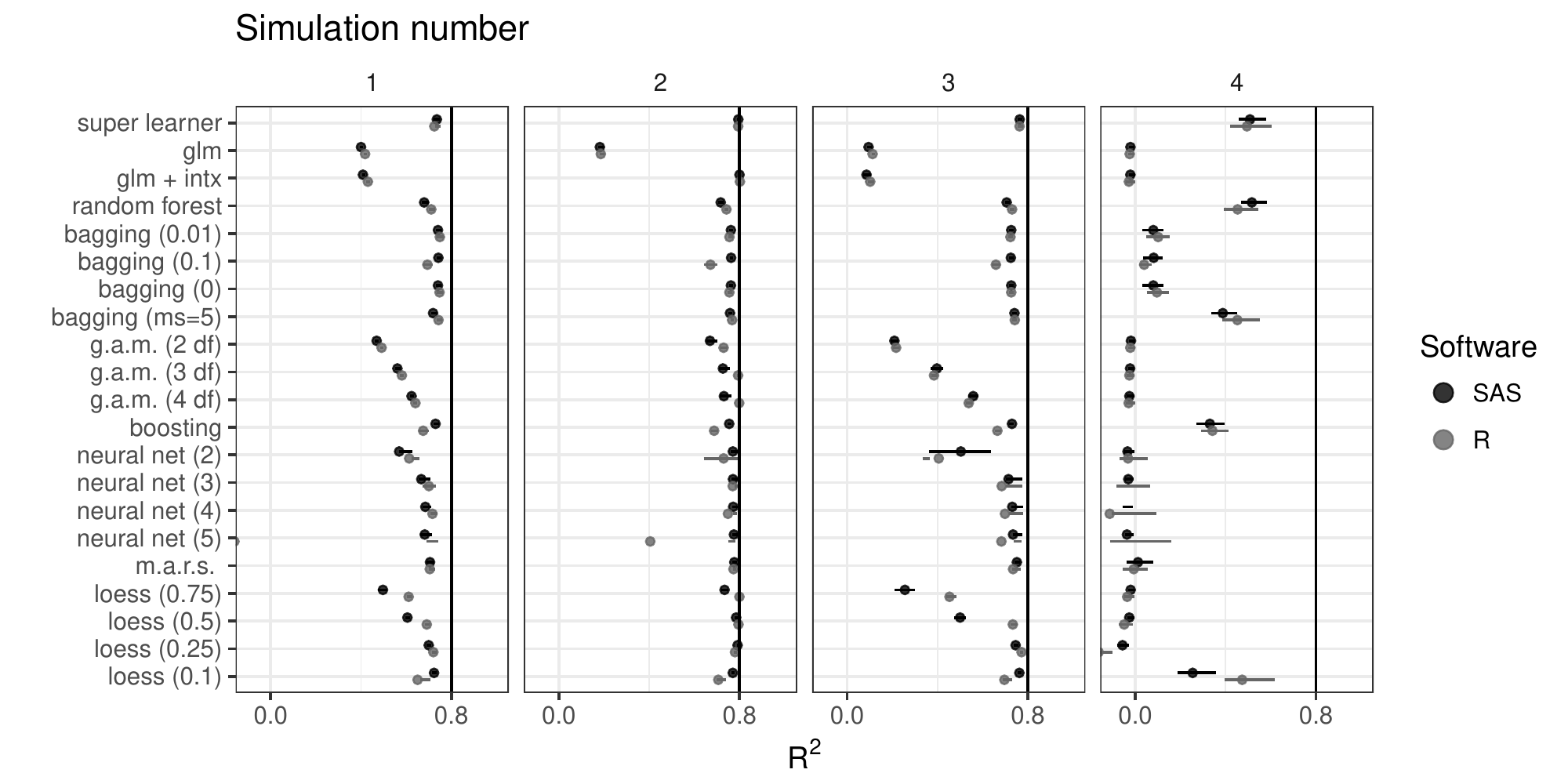}
\caption{\label{figsim1}Repeating Polley and van der Laan simulations \cite{RN566} assessing the out-of-sample prediction accuracy for super learner and algorithms in the super learner library.
The optimal value of $R^2$ is given by the black vertical line, and means are given with solid circles with interquartile ranges given by horizontal lines.}
\end{figure*}

An example call to the \SL{} macro for the analysis of the simulated data is given in Figure \ref{fig:code1}.
As shown in Figure \ref{figsim1}, super learner performed equally well in both R and SAS.
There were few meaningful differences across software platforms in the performance of individual learners, with the exception of \code{loess}, which is likely due to platform differences in smoothing kernel parameterization.
Notably, the SAS implementation demonstrated less variation with super learner predictions, likely due to the \code{neural net} and \code{bagging} algorithms which differed demonstrated higher variability in R.

\begin{figure}
\centering
\verbfilebox{snippets/call1}
 \fbox{\theverbbox}
 \caption{\label{fig:code1}Calling the \SL{} macro to carry out the simulation analysis described in section \ref{sec:sims}}
\end{figure}

\begin{figure}[h]
  \includegraphics[width=\linewidth]{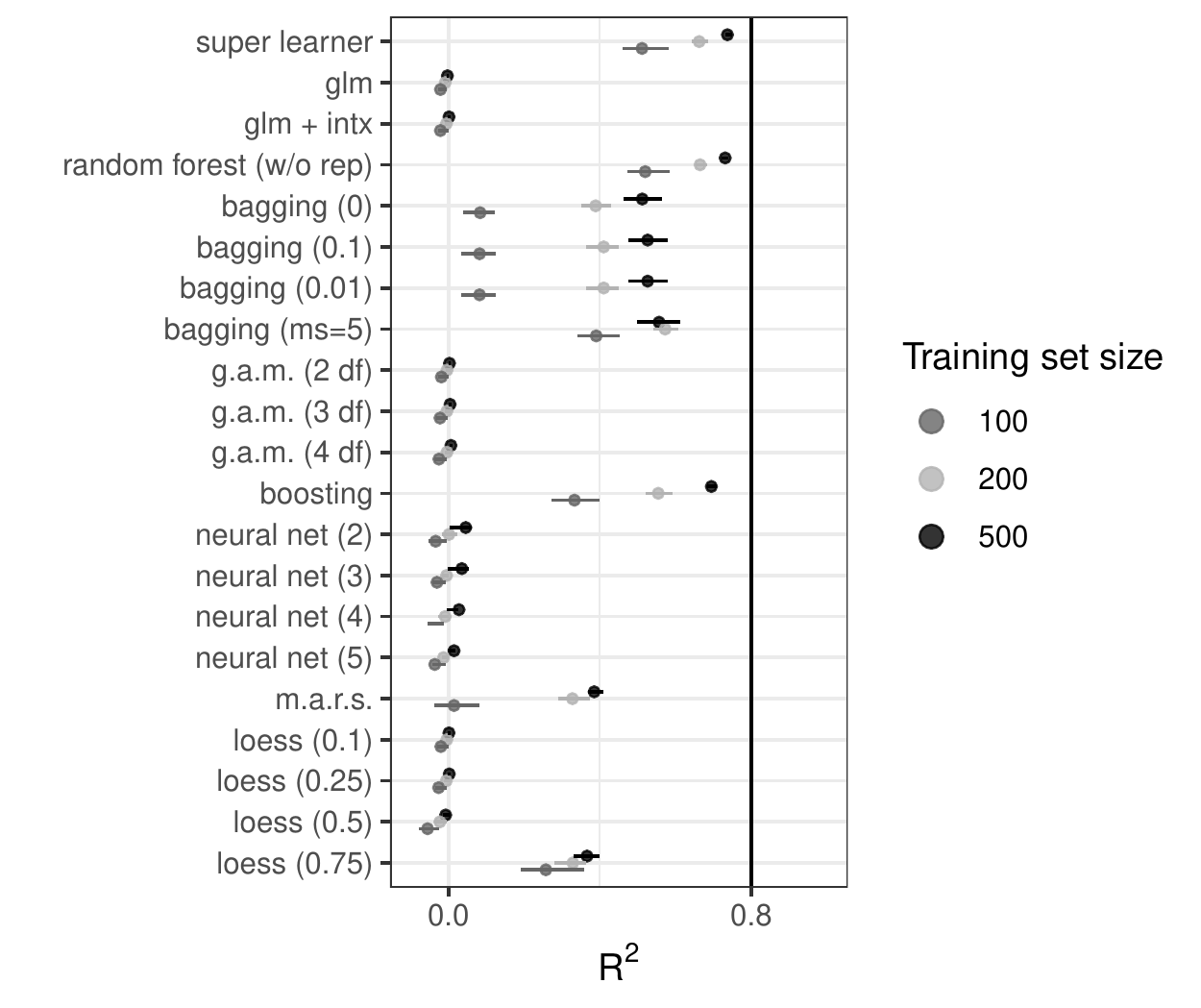}
\caption{\label{figsima1}Repeating simulation example 4 at varying sample sizes, demonstrating the benefit of additional training samples in super learner.}
\end{figure}

To improve $R^2$ estimates in Sim 4, Polley and van der Laan added to \code{super learner} a set of parametric models with basis functions for $x$ that included $sin$,  linear terms, and parameters for transitions between the two functions.
As an alternative to demonstrate how \code{super learner} can leverage data adaptive learners in the library, when the true data generating function is not even known approximately, we performed Sim 4 at additional training sample sizes of 200 and 500 to examine how each algorithm performed with more data.
Data adaptive algorithms such as \code{random forest}, \code{boosting},  \code{bagging}, \code{m.a.r.s.} and \code{loess} demonstrated improved estimation of $R^2$ at higher sample sizes, and \code{super learner} again performed as well as the best algorithm in the library at each sample size (Figure \ref{figsima1}).

\subsection{Example 2: Performance in 14 real world datasets \label{sec:real}}
To test the viability of using super learner in typical predictive problems, we tested the performance of the \SL{} macro in 14 real world data sets, closely following analyses reported in section 3.2 of Polley and van der Laan.
Each of the datasets had sample sizes between 200 and 700 (Table \ref{tab:tabreal}) and varied in the number of predictors from 3 to 20.
The datasets come from diverse subject matter areas including economic, health, engineering, and biologic data.
Some of the datasets used by Polley and van der Laan are no longer available, so only 10 of our 14 datasets were featured in Polley and van der Laan's analyses.

\begin{table}[h]
\caption{14 Real-world datasets used to evaluate super learning of a continuous variable in SAS and R}
\begin{tabular}{rccc}\hline
Data	&	N	&	p$^a$	&	Citation	\\\hline
ais	&	202	&	10	&	\citet{cook2009introduction}	\\
bodyfat	&	252	&	14	&	\citet{penrose1985generalized}	\\
cholesterol$^b$	&	297	&	13	&	\citet{james2013introduction}	\\
cps78	&	550	&	18	&	\citet{berndt1991practice}	\\
cps85	&	534	&	18	&	\citet{berndt1991practice}	\\
cpu	&	209	&	6	&	\citet{kibler1989instance}	\\
diabetes	&	375	&	15	&	\citet{harrell2001regression}	\\
diamond	&	308	&	17	&	\citet{chu2001pricing}	\\
fev	&	654	&	4	&	\citet{rosner2015fundamentals}	\\
house	&	506	&	13	&	\citet{harrison1978hedonic}	\\
mussels	&	201	&	3	&	\citet{cook2009regression}	\\
presidential$^b$	&	591	&	20	&	\citet{gelman2014bayesian}	\\
sat$^b$	&	339	&	4	&	\citet{carroll1997measurement}	\\
strike$^b$	&	625	&	6	&	\citet{western1996vague}	\\\hline
\end{tabular}\\
$^a$number of predictors\\
$^b$does not appear in \citet{RN566}.
\label{tab:tabreal}
\end{table}%

For each dataset, we defined one of the continuous variables as the target of interest ($Y$).
The objective of these analyses was to assess cross-validated predictive accuracy for $Y$ across all datasets for all candidate algorithms, including super learner.
We quantified predictive accuracy for each algorithm using 10-fold cross validated mean-squared error ($CVMSE_m$).
We followed Polley and van der Laan by scaling ${CVMSE_m}$ by the ${CVMSE_m}$ for the generalized linear model using 
$$\mbox{relative }MSE_m = CVMSE_m/CVMSE_{\mbox{glm}}.$$
To compare the average performance of each learner, we calculated the geometric mean of relative $MSE_m$ across all datasets for each learner.

To assess typical performance of super learning in SAS and R, we supplemented the library used in the simulation analyses of section \ref{sec:sims} to include a broader array of algorithms, including some that are limited to a single software package. We added \code{b spline} (basis splines, SAS only), \code{b.a.r.t.} (R only), \code{stepwise} (step-wise selection of a linear model \cite{Hastie:1992aa}), \code{ridge} (ridge regression \cite{Hastie:1992aa}), \code{l.a.s.s.o.}(least absolute shrinkage and selection operator \cite{Friedman:2010aa}), \code{bayes glm} (R only) \code{s.v.m} (support vector machine regression, R only \cite{Meyer:2017aa}), and \code{d.s.a.} (the deletion\slash substitution\slash addition algorithm \cite{Molinaro:2010aa}, R only).

\begin{figure}[h]
  \includegraphics[width=\linewidth]{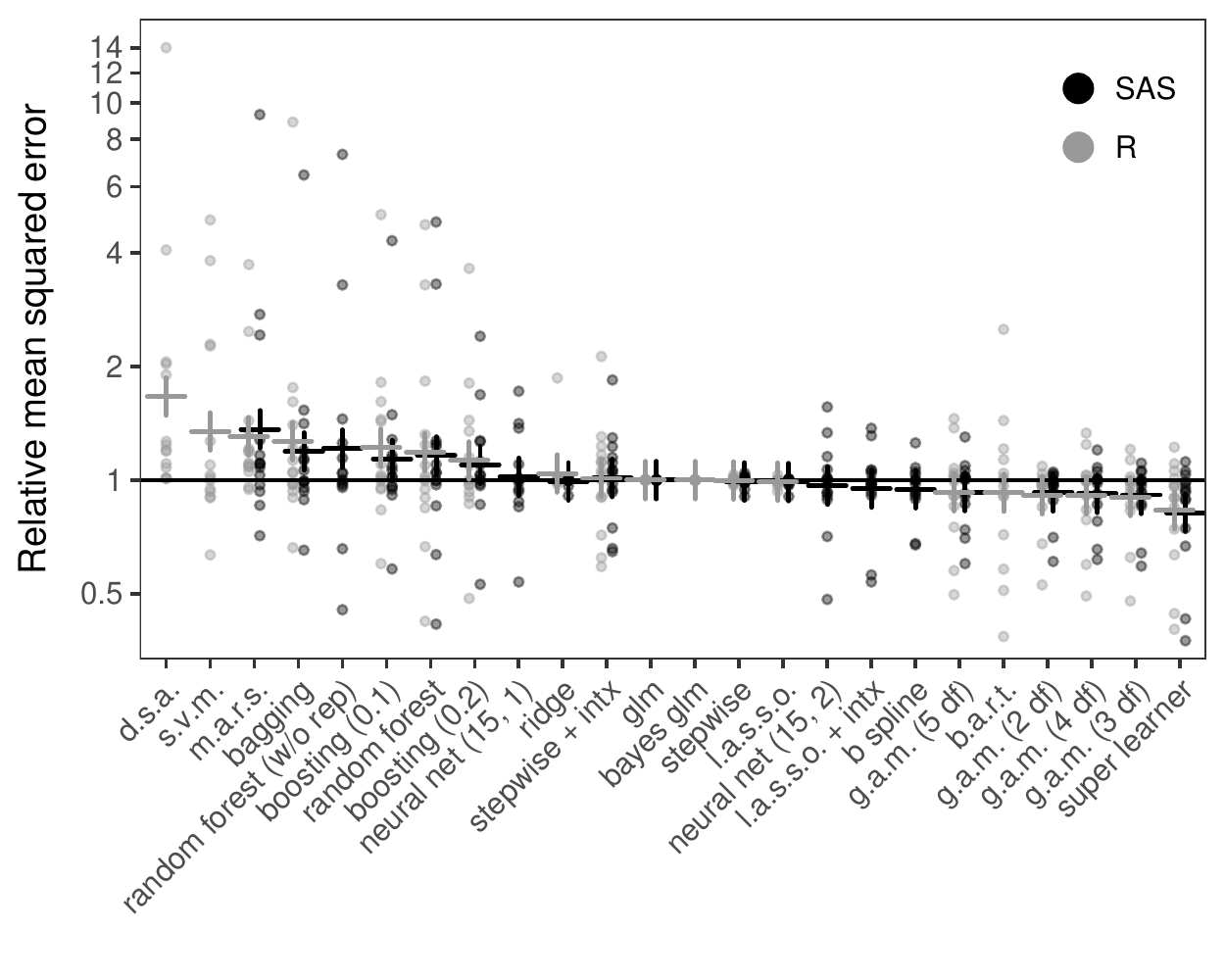}
\caption{\label{figreal1}Variant of Polley and van der Laan real data analysis \cite{RN566} assessing the 10-fold cross-validated relative mean-squared error (relative to \code{glm}) across 14 real datasets, sorted by geometric mean, denoted with a plus (+) sign.}
\end{figure}

In both SAS and R, the super learner algorithm had the lowest (best) average relative $MSE_m$ of all the algorithms examined (Figure \ref{figreal1}).
Among the library members, \code{g.a.m.} performed well under several different parameterizations.
Notably, the SAS version of \code{g.a.m.} demonstrated less variable performance relative to the R version.
In contrast with the findings of Polley and van der Laan, \code{b.a.r.t.} did not perform the best among the other algorithms, which may be due to differences across R packages or simply due to our use of different datasets.
Interestingly,  \code{m.a.r.s.} was more variable in SAS than in R.
The SAS and R versions of the \code{m.a.r.s.} differ in terms of default tuning parameter values, so the nominal category of the learner will not necessarily dictate its performance.
Some individual algorithmic differences aside, there appeared to be no important difference in average relative $MSE_m$ across these 14 datasets between Super Learner implemented with native-SAS procedures and Super Learner implemented with native-R packages.
Within a given dataset, however, they may not yield the same result due to differing performance among the available algorithms.

\section{Conclusions}

While the SuperLearner package has been available in R since 2010, there has been no such facility that is generally available in the SAS software system.
With the addition of the SAS Enterprise Miner software, the availability of machine learning algorithms in SAS has warranted a need for a way to combine inference from both parametric and data adaptive models.
The \SL{} macro fills this gap and, as demonstrated, performs similarly to the R package in a number of problems, even with a different set of natively available machine learning algorithms in each software system.

Aside from using the macro to explicitly make predictions using super learner, the macro provides a number of interesting benefits for analyses not based on super learning.
First, the macro can be used as a simple and automated way to access multiple of the machine learning features in SAS under a unified coding framework.
For example, one could use a nearly identical call to the macro to use both \code{random forest} and \code{b spline} even though the SAS syntax for the two underlying procedures (HPFOREST and TRANSREG) are very different.
Second, the macro provides a way to assess cross-validated measures of fit across multiple algorithms or model forms, simultaneously, thus automating an otherwise potentially tedious procedure that is not available by default in many SAS procedures.
Third, the macro could be explicitly used to select a software package for a given analysis: a call to the macro could include \code{random forest} implementations from both SAS and R. This is made possible through the RLANG option in SAS, which allows explicit calls to R algorithms (15 of which are included as default learners in the \SL{} macro).
Thus the macro provides a principled way to choose between R and SAS for a given problem.

The reliance on SAS places some constraints on the available features of the \SL{} macro.
Namely, while the \SL{} macro can be used to train super learner in one dataset in order to make predictions in another (as in the simulation example shown in figure \ref{fig:code1}), these processes must currently be done simultaneously.
The R package, on the other hand, allows one to save a trained super learner model for making predictions at a later time.
The procedural programming oriented approach of SAS make such a feature difficult to implement.
Further, many of the procedures underlying machine learning algorithms in SAS require the (paid) installation of SAS Enterprise Miner (e.g.
\code{random forest} and \code{neural net}) and even basic implementations require SAS/STAT and SAS/OR, which may or may not be included in some SAS installations.

Limitations notwithstanding, the \SL{} macro is a powerful tool that can draw on both the SAS and R systems for machine learning algorithms.
Thus, in the spirit of ensemble machine learning algorithms, this approach is appealing in the number of different learners that can be implemented as part of the super learner library.
This macro draws on a number of strengths from the SAS system, including the robust, enterprise oriented procedures, by-group processing, and the default capability to handle datasets that are too large to fit in memory.
Rather than replacing the SuperLearner package in R, this macro provides a valuable alternative to researchers more familiar with the SAS system or who use SAS due to enterprise features or collaborative ease.

\section*{Acknowledgements}
Funding for this work was provided through NIH/NICHD (Grant \# DP2-HD-08-4070) and other support was provided by the UNC Causal Inference Research Laboratory, which is supported by a Gillings Innovation Laboratory award.



\bibliography{ms.bbl}

\end{document}